\documentclass[10pt,twocolumn,letterpaper]{article}

\usepackage{iccv}
\usepackage{times}
\usepackage{epsfig}
\usepackage{graphicx}
\usepackage{amsmath}
\usepackage{amssymb}
\usepackage{upgreek,textgreek}

\usepackage{subfig}

\usepackage{tikz}
\usepackage{pgfplots}
\pgfplotsset{width=8.3cm,compat=newest}

\usepackage{tabularx}
\usepackage[T1]{fontenc}
\usepackage{booktabs}
\usepackage{siunitx}
\usepackage{lipsum}  

\usepackage[pagebackref=true,breaklinks=true,colorlinks,bookmarks=false]{hyperref}

\iccvfinalcopy 

\def\iccvPaperID{****} 

\usepackage{cuted}
\usepackage{capt-of}
\begin{document}

\title{Lumi\`ereNet: Lecture Video Synthesis from Audio}

\author{Byung-Hak Kim\thanks{Work done at Udacity, AI Team} 
\and Varun Ganapathi\footnotemark[1]}

\maketitle

\begin{strip}\centering
\includegraphics[width = 1.0\textwidth]{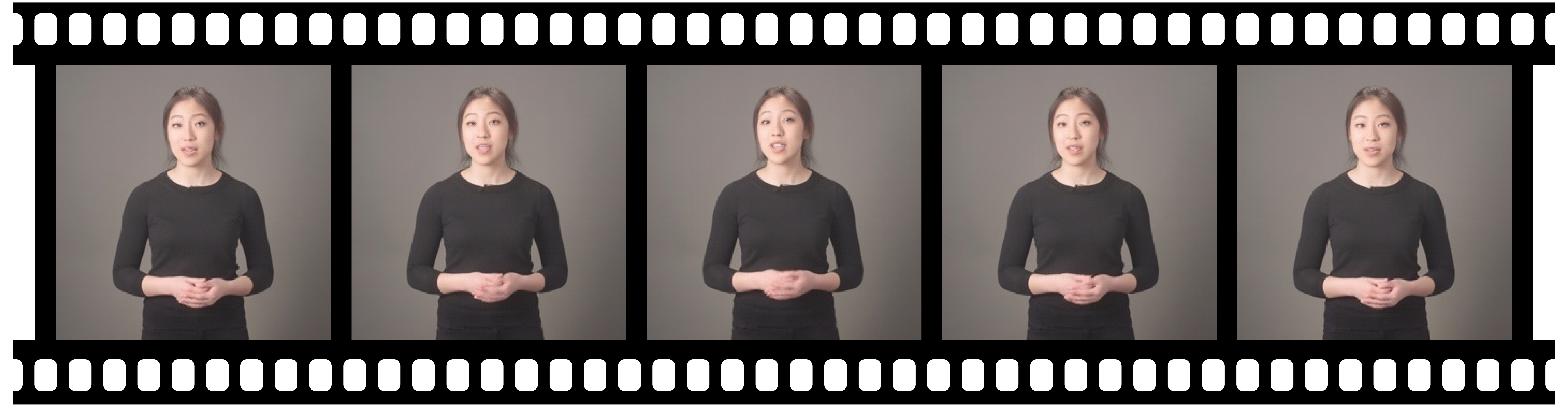}
\captionof{figure}{Synthesized full-pose headshot video of an instructor's lecture given her audio narration.\label{fig:teaser}}
\end{strip}

\begin{abstract}
   We present \textbf{Lumi\`ereNet}, a simple, modular, and completely deep-learning based architecture that synthesizes, high quality, full-pose headshot lecture videos from instructor's new audio narration of any length. Unlike prior works, Lumi\`ereNet is entirely composed of trainable neural network modules to learn mapping functions from the audio to video through (intermediate) estimated pose-based compact and abstract latent codes. Our video demos are available at \cite{Demo_new3:2019:Online} and \cite{Demo_new2:2019:Online}.
\end{abstract}

\section{Introduction}
\label{sec:introduction} 
To meet the increasing needs for people to keep learning throughout their careers, massive open online course (MOOCs) companies, such as Udacity and Coursera, not only aggressively
design new and relevant courses, but also frequently refresh existing course lectures to keep them up-to-date. In particular, instructor-produced lecture videos (i.e., a subject-matter expert delivering a lecture) are central in the current generation of MOOCs~\cite{Guo14,Hibbert14}. Due to their importance in online courses, video production is increasing exponentially, but production techniques are still not nimble enough to quickly shoot, edit, personalize, and internationalize the lecture video. This is because video production today requires considerable resources and processes (i.e., instructor, studio, equipment, and production staff) throughout the development phases. In current video production pipeline, an AI machinery which semi (or fully) automates lecture video production \emph{at scale} would be highly valuable to enable agile video content development (rather than re-shooting each new video). To that end, we propose a new method to synthesize lecture videos from any length of audio narration\footnote{We consider audio to video synthesis as a first step in this direction since audio narration recording is typically quicker and more affordable than video filming. This line of approach can be further pursued toward to a full lecture transcripts to video synthesis system.}. Given an instructor's lecture audios, we wish to synthesize corresponding video of any length. This problem of audio to video synthesis is, in general, challenging, as we must learn a mapping which goes from lower dimensional signals (audio) to a higher dimensional (3D) time-varying image sequences (videos). However, with the availability of video stock footage of a subject teaching and significant recent advances in the community, we are able to discover this mapping between audio and corresponding visuals directly in a supervised way.  

\begin{figure*}[ht]
\centering
\vspace{.5cm}
\usetikzlibrary{shapes.geometric, shapes.multipart, arrows, calc, positioning}

\tikzstyle{blstm} = [rectangle, rounded corners, minimum width=1.5cm, minimum height=1.25cm,text centered, draw=black,fill=red!30]
\tikzstyle{dec} = [trapezium, rounded corners, minimum width=1.25cm, minimum height=1cm,shape border rotate=90,text centered, draw=black, fill=blue!30]
\tikzstyle{gan} = [rectangle, rounded corners, minimum width=1.5cm, minimum height=1.25cm,text centered, draw=black, fill=green!40]

\tikzstyle{arrow} = [thick,->,>=stealth]

\begin{tikzpicture}[node distance=2.0cm]

    \tikzstyle{every node}=[font=\small]
    
    \node[align=center] (in_feat) {Audio\\features};
    \node[align=center] (fw0) [blstm, right of=in_feat, xshift=0.8cm] {BLSTM\\model};
    \node[align=center] (fw1) [dec, right of=fw0, xshift=0.8cm] {VAE\\decoder};
    \node[align=center] (fw2) [gan, right of=fw1, xshift=0.8cm] {SeqPix2Pix\\model};
    \node[align=center](out_feat)[right of=fw2, xshift=0.8cm] {Video\\frames};

    \draw[arrow] (in_feat) -- (fw0) node[midway,above] {\(\mathbf{x}\)};
    \draw[arrow] (fw0) -- (fw1) node[midway,above] {\(\mathbf{z}\)};
    \draw[arrow] (fw1) -- (fw2) node[midway,above] {\(\mathbf{w}\)};
    \draw[arrow] (fw2) -- (out_feat) node[midway,above] {\(\mathbf{y}\)};
    
\end{tikzpicture}
\vspace{1cm}
\input{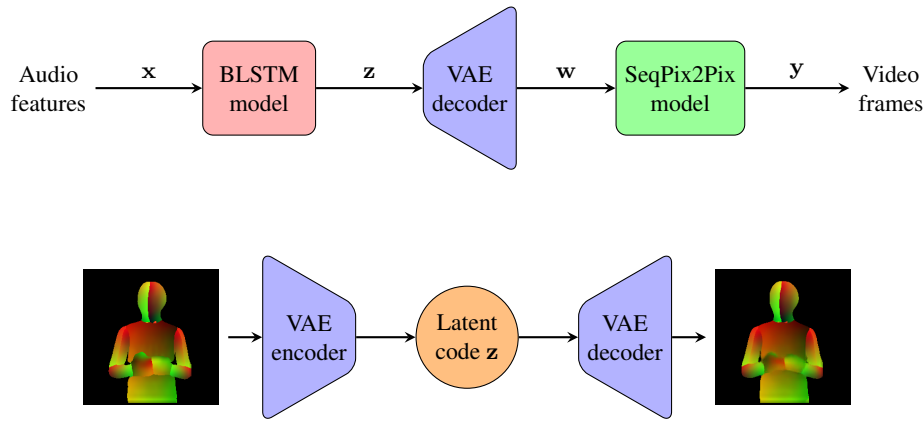}
\vspace{.5cm}
\caption{Proposed Lumi\`ereNet architecture overview. Lumi\`ereNet consists of three neural network modules: the \textbf{BLSTM model}, the \textbf{VAE model}, and the \textbf{SeqPix2Pix model}. The BLSTM model first associates extracted audio features \(\mathbf{x}\) to intermediate latent codes \(\mathbf{z}\). Then, the VAE decoder constructs the corresponding pose figures \(\mathbf{w}\) from \(\mathbf{z}\). Lastly, the SeqPix2Pix model produces the final video frames \(\mathbf{y}\) given \(\mathbf{w}\). During training, Lumi\`ereNet learns the VAE model to design compact and abstract latent codes \(\mathbf{z}\) for high-dimensional DensePose images altogether with both encoder and decoder.}
\label{LumereNetArch}
\end{figure*}

There have been several important attempts in this direction by focusing on synthesizing parts of the face (around the mouth)~\cite{Suwajanakorn17,Kumar17}. However, as instructors' emotional states are communicated not only with facial expressions, but also through body posture, movement, and gestures~\cite{Dael12,Lhommet15}, we introduce a pose estimation based latent representation as an intermediate \emph{code} to synthesize an instructor's face, body, and the background altogether. We design these compact and abstract codes from the extracted human poses for a subject which allows video image frame and audio sequences to be conditionally independent given them. It is convenient to think of the obtained pose detection that each frame yields as a corresponding set of \big(audio, pose figures) and \big(pose figures, person images). 

Our primary contributions are twofold. We present a fully neural network-based, modular framework which is sufficient to achieve convincing video results. This framework is simpler than prior classic computer vision based models. We also illustrate the effects of several important architectural choices for each sub-module network. Even though our approach is developed with primary intents to support agile video content development which is crucial in current online MOOC courses, we acknowledge there could be potential misuse of the technologies. Nonetheless, we believe it is crucial synthesized video using our approach requires to indicate as synthetic and it is also imperative to obtain consent from the instructors across the entire production processes.

\section{Related Work}
\label{sec:related work}
\subsubsection*{Visual speech synthesis}
\label{ssec:visual speech synthesis}
Over the last two decades, there has been extensive study dedicated towards creating realistic animations for speech \cite{Thies16} in 2D or 3D. 2D has the advantage that video cutouts of the mouth area can be used and combined leading to realistic visualizations \cite{Bregler97,Ezzat97}. 3D approaches are much more versatile, as viewpoints and illumination can be changed at will \cite{Karras17}. Given that our goal is to produce 2D animation based on audio, instead of formulating entire mapping as an end-to-end optimization task, we are inherently interested in the intermediate representations. Recent advances in this line were more or less focused on synthesizing only the parts of the face (around the mouth) and borrowing the rest of the subject from existing video footage~\cite{Shimba15,Suwajanakorn17,Kumar17}.  

\subsubsection*{Human pose estimation}
\label{ssec:human pose estimation}
Human pose estimation is a general problem in computer vision to detect human figures in images and video. Recent deep-learning based algorithmic advances enable not only the detection and localization of major body points, but detailed surface-based human body representation (e.g.,OpenPose~\cite{Cao17} and DensePose~\cite{Guler18}). We use a pre-trained DensePose estimator to create body figure RGB images from video frames.  

\subsubsection*{Image-to-Image translation}
Several recent frameworks have used generative adversarial networks (GANs)~\cite{Goodfellow14} to learn a parametric translation function between input and output images \cite{Isola17}. Similar ideas have been applied to various tasks, such as generating photographs from sketches, or even video applications for both paired and unpaired cases~\cite{Zhu17,Chan18,Wang18,Bansal18}. However, none of these techniques are fool-proof, and some amount of limitations often remain.\\

\section{Lumi\`ereNet}
\label{sec:LumiereNet}
\subsection{Problem Formulation}
\label{ssec:problem formulation}
We consider the problem of synthesizing a video from an audio expressed as ``How to learn a function to map from an audio sequence \(\mathbf{x}\triangleq(x_{1},\dots,x_{T})\) recorded by an instructor to video frame sequence \(\mathbf{y}\triangleq(y_{1},\dots,y_{T'})\)?''\footnote{For the sake of brevity, we assume both audio and video frame sequences as discrete-time sequences.}. Answers to this question certainly require some assumptions about the video generation process, so we begin differently by introducing a probabilistic model for video generation. The basic idea is that two hidden (or intermediate) representations of pose-estimation \(\mathbf{w}\) 
and corresponding compact and abstract codes \(\mathbf{z}\) defined as
\begin{equation}
\mathbf{w}\triangleq(w_{1},\dots,w_{T'}), 
\,\mathbf{z}\triangleq(z_{1},\dots,z_{T'}) \\
\end{equation}
are introduced for audios and videos. This allows that the video frames \(\mathbf{y}\) and audio sequences \(\mathbf{x}\) to be conditionally independent given \(\mathbf{w}\) (i.e., \(P(\mathbf{y}|\mathbf{w},\mathbf{x})\triangleq P(\mathbf{y}|\mathbf{w})\)). Additionally, \(\mathbf{x}\) and \(\mathbf{w}\) are also conditionally independent given \(\mathbf{z}\) (i.e., \(P(\mathbf{w}|\mathbf{z},\mathbf{x})\triangleq P(\mathbf{w}|\mathbf{z})\)). With the goal to design a probabilistic mapping that reflects entire associations, the conditional independence assumptions in the model imply that
\begin{equation} \label{eq1}
\begin{split}
P(\mathbf{y}|\mathbf{x}) & \triangleq P(\mathbf{y}|\mathbf{w})P(\mathbf{w}|\mathbf{z})P(\mathbf{z}|\mathbf{x}). \\
\end{split}
\end{equation}

Specifically, we consider neural network modules as a randomly chosen instance of this problem based on this \emph{probabilistic generative} model. A key advantage is that each neural network model represents a single factor which separates the influence of other networks that can be trained and improved independently. This also greatly reduces each network's complexity and size as the dimension of compact latent codes \(\mathbf{z}\) (128 in our experiments) is typically much smaller than \(\mathbf{w}\). In fact, this model can be seen as a generalization of simpler models in \cite{Suwajanakorn17} and \cite{Kumar17}. 

\subsection{Designing \(P(\mathbf{w}|\mathbf{z})\)}
\label{ssec:VAE}
\subsubsection*{DensePose estimator} 
We use a pretrained DensePose system~\cite{Guler18} to construct human pose figures \(\mathbf{w}\) of video frame sequences \(\mathbf{y}\). Even though DensePose results do not account for fine details explicitly (like eye motion, blink, lip, hair, clothing), structural poses are largely well captured. Of course, DensePose estimation correctness can be compromised by inaccuracies of the system and self-occlusions of the body\footnote{We do not account for these occlusions as most lecture video shots are relatively controlled with the subject in the center and facing the camera.}. 

\subsubsection*{VAE model} 
The role of the variational auto encoder (VAE) model is to learn a compressed abstract representation of each estimated DensePose image. Here, one could use a simple model~\cite{Kingma14} as our VAE model to encode each high-dimensional DensePose image \(w_{i}\) into a low-dimensional latent code \(z_{i}\) respectively. While the main role of the VAE model is to squeeze each DensePose image, we also want to have reconstructed pose figures to have better perceptual quality. To that end, instead of using classic pixel-by-pixel loss, we use ImageNet-pretrained VGG-19 \cite{Ronneberger15} based perceptual loss\footnote{We compute Euclidean distance between input and output images after projecting them into abstract feature spaces using the pool1, pool2 and pool3 layers of VGG-19.}, which has shown previously to help the VAE's output to preserve spatial correlation characteristics of the input image~\cite{Hou17}. Given this VAE model, the decoder part which produces \(\mathbf{w}\) for latent codes \(\mathbf{z}\) is used as \(P(\mathbf{w}|\mathbf{z})\).

\subsection{Mapping for \(P(\mathbf{z}|\mathbf{x})\)}
\label{ssec:BLSTM}
\subsubsection*{Audio features extractor}
We represent audio signals using the log Mel-filterbank energy features. In the log filterbank energy computations, we apply 40 filters of 1024 size with 44ms-length sliding window at a 33.3ms sampling interval. This configuration is a simple way to match the final video generation rate (30 Hz in our experiments), while keeping frame shift to be 75\% of the frame size without adding complications (c.f., upsampling DensePose images). Lastly, we apply normalization to the audio features to feed them to the bidirectional long short term memory (BLSTM) based neural networks~\cite{Graves05} described below.

\subsubsection*{BLSTM model} 
With the VAE model's encoder output \(\mathbf{z}\), we try to learn a mapping that associates encoded \(\mathbf{z}\) with the audio features sequence \(\mathbf{x}\). To ensure the input audios to be well-aligned (or conditioned) with the outputs, we employ future and past audio contexts via a single BLSTM layer concatenating forward and backward direction LSTM outputs, followed by a linear fully-connected (FC) layer of a code dimension. Also, to help ensure the outputs are coherent over time without abrupt changes or jumps, we prepare each input to have its own look-back window of length \(W\). One could potentially consider learning to map directly to the pixel-level DensePose image space (not to latent embedding space). However, this would add high redundancies in the output layer and require much greater memory. 

\subsection{Mapping for \(P(\mathbf{y}|\mathbf{w})\)}
\label{ssec:sequential pix2pix}
\subsubsection*{SeqPix2Pix model}
Our SeqPix2Pix model builds on the Pix2Pix framework \cite{Isola17}, which uses a conditional generative adversarial network~\cite{Goodfellow14} to learn a mapping from input to output images. There have been a few attempts to overcome limitations of the application of conventional Pix2Pix algorithms to video settings. Most research focused on the fact that \(\mathbf{y}\) consists of temporally ordered streams, so a model has to produce not only photorealistic, but spatio-temporally coherent frames as well~\cite{Chan18,Wang18}. 

The classic Pix2Pix formulation can be described as learning a mapping \(G: \mathbf{w} \rightarrow{}\mathbf{y}\), for each \(t\), as 
\begin{equation} \label{eq2}
\begin{split}
\min_{G}\max_{D} L_{t}(G,D)\\
\end{split}
\end{equation}
where \(L_{t}(G,D)\triangleq\log D(y_{t})+\log(1-D(G(w_{t}))\).
The key assumption of this approach is synthesizing each image independently across time as:
\begin{equation} 
\begin{split}
P(\mathbf{y}|\mathbf{w}) & \triangleq \prod_{t=1}^{T'} P(y_{t}|w_{t}).\\
\end{split}
\end{equation}
This would potentially limit the capabilities of temporal smoothing in output generations. Even though \(P(\mathbf{z}|\mathbf{x})\) produces output streams that naturally exhibit temporal continuity, it would be beneficial to have additional temporal smoothness in the \(P(\mathbf{y}|\mathbf{w})\) model. 

One way to reflect the memory property of the video sequence is by incorporating the following Markov property with memory length \(L\):
\begin{equation} \label{eq3}
\begin{split}
P(\mathbf{y}|\mathbf{w}) & \triangleq \prod_{t=1}^{T'} P(y_{t}|y_{t-L},\dots,y_{t-1};w_{t-L},\dots,w_{t}).\\
\end{split}
\end{equation}
With this new assumption, we extend the Pix2Pix framework into a sequential setting by introducing a temporal predictor \(P:(y_{t-L},\dots,y_{t-1}) \rightarrow{} y_{t}\)\footnote{The idea of temporal predictor was inspired from~\cite{Bansal18}.}: 
\begin{align}
  \min_{(G,P)}\max_{D} &\null  \,L_{t}(G,D,P) \label{eq4}\\
  \text{subject to } &\null y_{t}=G(w_{t}), \label{eq5}\\
  &\null y_{t}=P(G(w_{t-L}),\dots,G(w_{t-1})), \label{eq6}\\
  &\null y_{t}=P(y_{t-L},\dots,y_{t-1}) \label{eq7}, 
\end{align}
where \(L_{t}(G,D,P)\) is a modifed GAN loss over memory length \(L\) defined as \(\sum_{i=t-L}^{t} \log D(y_{i})+\log(1-D(G(w_{i}))\). 

Each additional constraint from Equation~\ref{eq5} to Equation~\ref{eq7} has different purposes. First Equation~\ref{eq5} is what we call structural consistency constraint, and the last two constraints Equation~\ref{eq6} and ~\ref{eq7} are temporal consistency constraints among the last \(L\) samples. In fact, these added constraints act as barrier functions to guide the convergence to better local optima. We can rewrite the final SeqPix2Pix formulation as:
\begin{align}
  \min_{(G,P)}\max_{D}\,&\null  L_{t}(G,D,P) \\
  +\,&\null \lambda_{0}\,l_{L_{2}}(\Phi(y_{t}),\Phi(G(w_{t}))) \\
  +\,&\null \lambda_{1}\,l_{L_{1}}(y_{t}, P(G(w_{t-L}),\dots,G(w_{t-1}))) \\
  +\,&\null \lambda_{2}\,l_{L_{1}}(y_{t}, P(y_{t-L},\dots,y_{t-1})) 
\end{align}
where \(l_{L_{2}}\) is the Euclidean distance between feature representations \(\Phi(\cdot)\)\footnote{Again, the pool1, pool2, and pool3 layers of VGG-19 are used for our experiments.}~\cite{Johnson16}, and \(l_{L_{1}}\) is the \(L_{1}\) distance between pixel-level images. Unlike the original Pix2Pix, we opt out of the PatchGAN discriminator \cite{Isola17} because global structural properties are largely settled in \(\mathbf{w}\). 


\section{Experiments}
\label{sec:experiment}
\subsection{Video shootings}
\label{ssec:Dataset construction}
We filmed our instructor's lecture video for around 4 hours\footnote{Unlike prior works \cite{Suwajanakorn17,Kumar17} which relied solely on male's video footage, we decided to shoot a female instructor in order to enhance diversity in related literature.}. The footage was filmed in an in-house studio at Udacity and used the same setup as a regular production shoot. The studio had plain grey paper backgrounds, lights, an iPad prompter, and a single close-up C100 camera. Lecture transcripts are all prepared by the instructor, which are broken up into chunks that are about the same length of 3 to 4 minutes. We had the instructor read from the prompter. We kept a higher error tolerance than during regular shooting. In other words, we didn't stop for mistakes or have the instructor reshoot to fix errors, due to time constraints. Instead, we asked to continue naturally, even when there were errors with the instructor's delivery. This allowed us to complete filiming without making mistakes too obvious. For usual production shooting, we would do at least two retakes, or have the instructor continue until the take is perfect. With these guidelines, there was essentially one take of each script, and for four hours of video, it took about 8 hours of shooting.

\subsection{Data prepossessing}
\label{sec:data preprocessing}
Our instructor's videos were filmed at 30 frames per second. Each video is about 3 to 4 minutes and 1920x1080 resolution. We resized each frame to 455x256 resolution and cropped the central 256x256 regions. Audios are extracted from videos and converted from 48kHz to 16kHz. We use a ResNet-101 based DensePose estimator trained on cropped person instances from the COCO-DensePose dataset. The output consists of 2D fields representing body segments and \({U,V}\) coordinate spaces aligned with each of the semantic parts of the 3D model. For SeqPix2Pix model training, we used 1 image for every 30 frames. Unlike other works, we did not apply any other manual prepossessing.

\subsection{Network architectures}
\subsubsection*{VAE model}
Both the encoder and decoder networks are based on convolutional neural networks. We use three convolutional layers in the encoder network with 3x3 kernels. Each convolutional layer is followed by a ReLU activation layer and maxpooling layer of size 2. Then two fully-connected output layers for mean and variance are added to the encoder. For decoder network, we use 4 convolutional layers with 3x3 kernels with  upsampling each layer by a factor of 2. We also use ReLU as the activaion function.

\subsubsection*{BLSTM model}
The BLSTM model is composed of forward and backwards LSTM layers containing 256 cell dimensions per direction and followed by an FC layer of 128 dimension. We set the look-back window length \(W = 15\).

\subsubsection*{SeqPix2Pix model}
We use the U-Net based Pix2Pix generative network architecture for \(G\) \cite{Ronneberger15} and the DCGAN discriminator \cite{Radford16} for \(D\). The temporal predictor \(P\) concatenates the last \(L\) frames (2 for our experiments) as an input to the identical U-Net architecture to predict next frame. 

\subsection{Training details}
\label{ssec:training}
We first train the VAE model to encode DensePose frames into latent \(\mathbf{z}\) space. The VAE model is trained using an RMSProp optimizer \cite{Hinton12} with \(lr = 0.00025\). After that, we train the BLSTM and SeqPix2Pix models in parallel. The BLSTM model is trained for \(L_{2}\) loss using an RMSProp optimizer with \(lr = 0.000001\). For the SeqPix2Pix model, we replace the negative log likelihood in modified GAN loss by a least-squares loss~\cite{Mao17}, set \((\lambda_{0},\lambda_{1},\lambda_{2})=(0.05,10.0,10.0)\), and use the ADAM \cite{Kingma15} optimizer with \(lr = 0.0002\) and \((\beta_{1},\beta_{2})=(0.5,0.999)\).

\begin{figure}[t]
\captionsetup{position=top}
\captionsetup[subfigure]{labelformat=empty}
\begin{center}
\subfloat[Original Frame \hspace{15mm} Reconstructed Frame]{\includegraphics[width = 0.4\textwidth]{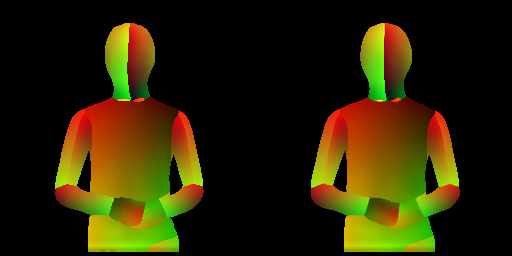}}\\[-3.ex]
\subfloat{\includegraphics[width = 0.4\textwidth]{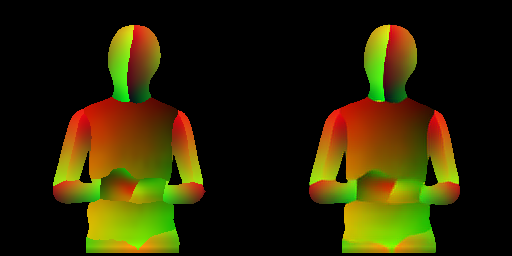}}\\[-3.ex] 
\subfloat{\includegraphics[width = 0.4\textwidth]{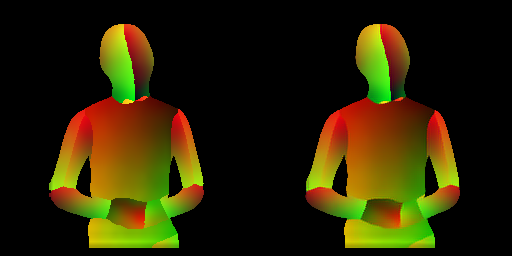}}\\[-3.ex] 
\subfloat{\includegraphics[width = 0.4\textwidth]{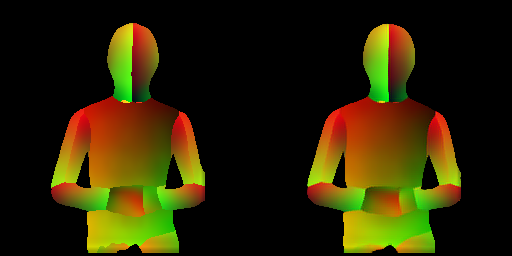}}\\[-3.ex]
\subfloat{\includegraphics[width = 0.4\textwidth]{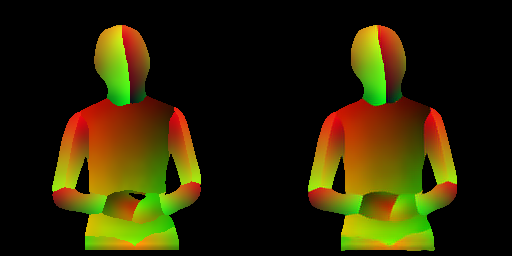}}\\[-3.ex]
\end{center}
\caption{VAE reconstruction results. We show five frames in sequential order across nine seconds and compare to the original DensePose frames.}
\label{fig:VAE exmample}
\end{figure}

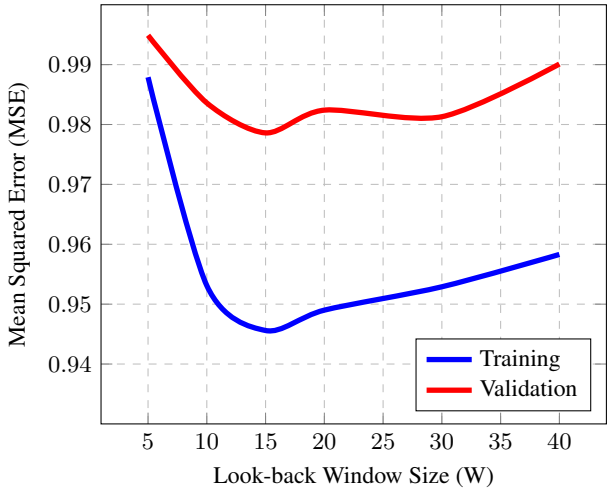
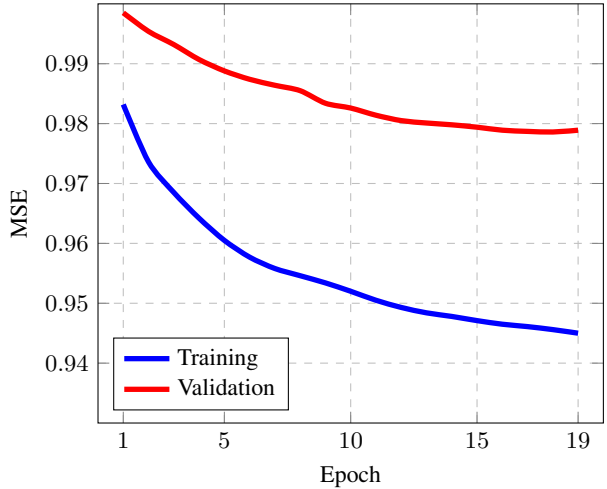
\begin{figure*}[ht]
\centering
\subfloat[Training and validation losses for various look-back window sizes\label{fig:BLSTMperformance_a}]{\begin{tikzpicture}

\begin{axis}[
    xlabel={Look-back Window Size (W)}, 
    ylabel={Mean Squared Error (MSE)},
    xmin=1, xmax=44, xtick={5,10,...,35,40},
    ymin=0.93, ymax=1.0, ytick={.94,.95,.96,.97,.98,.99},
    xmajorgrids=true, 
    ymajorgrids=true, 
    grid style=dashed,
    label style={font=\small}, 
    tick label style={font=\small},  
    legend pos=south east,
    legend style={font=\small},
    legend cell align={left}
]

\addplot[color=blue,line width=2.0pt,smooth]
    coordinates {
                (5,.9879)
                (10,.9531)
                (15,.9456)
                (20,.9490)
                (30,.9529)
                (40,.9583)};

\addplot[color=red,line width=2.0pt,smooth]
    coordinates {
                (5,.9949)
                (10,.9836)
                (15,.9786)
                (20,.9824)
                (30,.9813)
                (40,.9901)};

\legend{Training, Validation}
\end{axis}

\end{tikzpicture}}
\hfill
\subfloat[The losses during training of the BLSTM model with \(W^{*}\)=15\label{fig:BLSTMperformance_b}]{\begin{tikzpicture}

\begin{axis}[
    xlabel={Epoch}, 
    ylabel={MSE},
    xmin=0, xmax=20, xtick={1,5,10,...,15,19},
    ymin=0.93, ymax=1.0, ytick={.94,.95,.96,.97,.98,.99},
    xmajorgrids=true, 
    ymajorgrids=true, 
    grid style=dashed,
    label style={font=\small}, 
    tick label style={font=\small},  
    legend pos=south west,
    legend style={font=\small},
    legend cell align={left}
]

\addplot[color=blue,line width=2.0pt,smooth]
    coordinates {
                (1,0.9832)
                (2,0.9736)
                (3,0.9685)
                (4,0.9642)
                (5,0.9605)
                (6,0.9577)
                (7,0.9558)
                (8,0.9546)
                (9,0.9534)
                (10,0.9520)
                (11,0.9505)
                (12,0.9493)
                (13,0.9484)
                (14,0.9478)
                (15,0.9471)
                (16,0.9465)
                (17,0.9461)
                (18,0.9456)
                (19,0.9450)
                };

\addplot[color=red,line width=2.0pt,smooth]
    coordinates {
                (1,0.9985)
                (2,0.9954)
                (3,0.9932)
                (4,0.9907)
                (5,0.9888)
                (6,0.9874)
                (7,0.9864)
                (8,0.9855)
                (9,0.9834)
                (10,0.9826)
                (11,0.9814)
                (12,0.9805)
                (13,0.9801)
                (14,0.9798)
                (15,0.9794)
                (16,0.9789)
                (17,0.9787)
                (18,0.9786)
                (19,0.9789)
                };

\legend{Training, Validation}
\end{axis}

\end{tikzpicture}}
\hfill
\caption{Loss curves of the BLSTM model on different look-back window sizes. (a): Measured MSE losses at the end of the training of the same BLSTM model with varying look-back window length. (b): Reported MSE losses during training for the BLSTM model with \(W^{*}\)=15.}
\label{fig:BLSTMperformance}
\end{figure*}

\newcolumntype{C}{>{\raggedleft\arraybackslash}X} 
\begin{table*}[h]
  \caption{Quantitative results on images from Test Set 1 and Test Set 2. We report mean MSE / PSNR / SSIM of each model for each test dataset.}
  \label{table: Qualitative results}
  \centering
  \begin{tabularx}{0.8\textwidth}{@{}l*{11}{C}c@{}}
    \textbf{Ground  Truth} & \textbf{Baseline 1} & \textbf{Baseline2} & \textbf{SeqPix2Pix} \\ 
    \midrule
    Test Set 1 mean & 114.78 / 27.64 / 0.91 & 80.68 / 29.16 / 0.93 & 78.97 / 29.25 / 0.88 \\ 
    Test Set 2 mean & 116.51 / 27.61 / 0.91 & 77.06 / 29.37 / 0.93 & 77.39 / 29.35 / 0.88 \\ 
  \end{tabularx}
\end{table*}

\subsection{Experiment Results}
\label{ssec:perceptual assessments}
\subsubsection*{VAE model}
We show several qualitative example results in Figure~\ref{fig:VAE exmample}. Overall, the VAE model is able to reconstruct an almost perfect image when the instructor's face is square, facing directly at the camera. Body shapes are very well translated. The hands look good, but on closer examination, the lines between the two hands' fingers often look blurry and would be one of major causes for poor final generation outputs. 

\subsubsection*{BLSTM model}
\label{sssec:BLSTM model}
We learned that preparing each audio feature input to have its own look-back window is crucial for improving validation loss and visual consistency of the reconstructed pose figures (generated by the VAE decoder). Figure~\ref{fig:BLSTMperformance}\subref{fig:BLSTMperformance_a} shows training and validation losses for different look-back window sizes for the same BLSTM model (i.e., forward and backwards LSTM layers containing 256 cell dimensions per direction and followed by an FC layer of 128 dimension). Figure~\ref{fig:BLSTMperformance}\subref{fig:BLSTMperformance_b} plots the losses during training of the BLSTM model with look-back window length \(W^{*}\)=15, which used in our experiments.

\subsubsection*{SeqPix2Pix model}
\begin{figure*}[ht]
\captionsetup{position=top}
\captionsetup[subfigure]{labelformat=empty}
\begin{center}
\subfloat[Baseline 1]{\includegraphics[width = 0.22\textwidth]{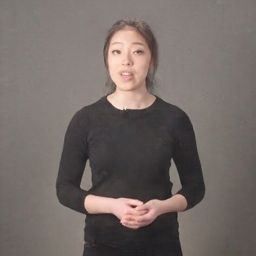}}
\subfloat[Baseline 2]{\includegraphics[width = 0.22\textwidth]{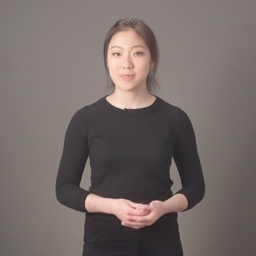}}
\subfloat[SeqPix2Pix]{\includegraphics[width = 0.22\textwidth]{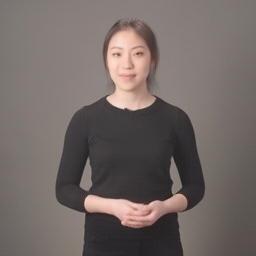}}
\subfloat[Ground Truth]{\includegraphics[width = 0.22\textwidth]{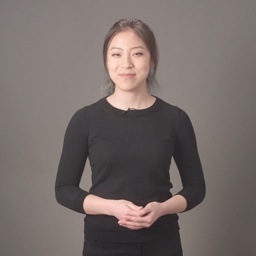}}\\[-2.5ex]
\subfloat{\includegraphics[width = 0.22\textwidth]{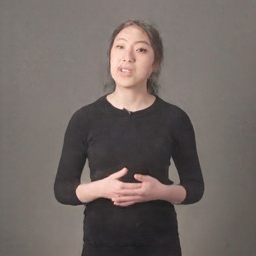}}
\subfloat{\includegraphics[width = 0.22\textwidth]{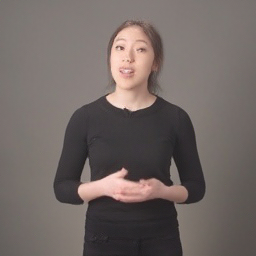}}
\subfloat{\includegraphics[width = 0.22\textwidth]{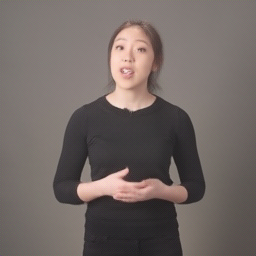}}
\subfloat{\includegraphics[width = 0.22\textwidth]{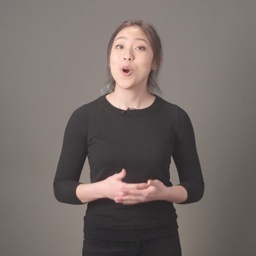}}\\[-2.5ex]
\subfloat{\includegraphics[width = 0.22\textwidth]{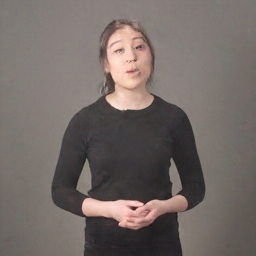}}
\subfloat{\includegraphics[width = 0.22\textwidth]{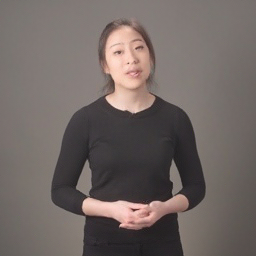}}
\subfloat{\includegraphics[width = 0.22\textwidth]{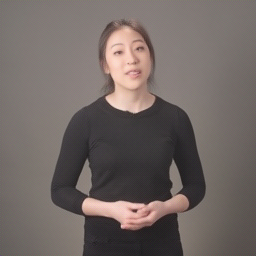}}
\subfloat{\includegraphics[width = 0.22\textwidth]{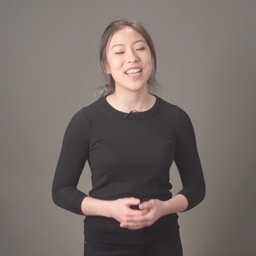}}\\[-2.5ex]
\subfloat{\includegraphics[width = 0.22\textwidth]{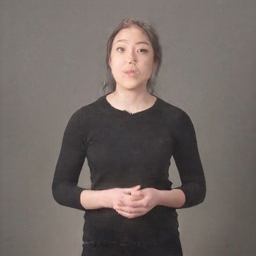}}
\subfloat{\includegraphics[width = 0.22\textwidth]{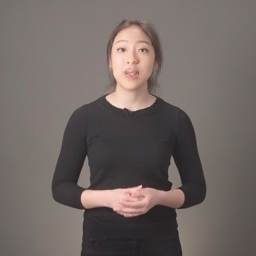}}
\subfloat{\includegraphics[width = 0.22\textwidth]{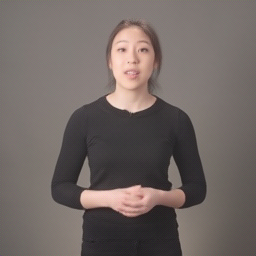}}
\subfloat{\includegraphics[width = 0.22\textwidth]{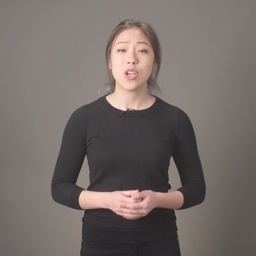}}\\[-2.5ex]
\subfloat{\includegraphics[width = 0.22\textwidth]{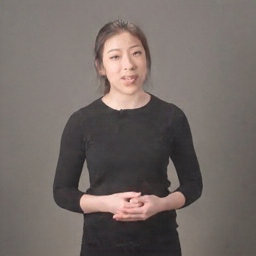}}
\subfloat{\includegraphics[width = 0.22\textwidth]{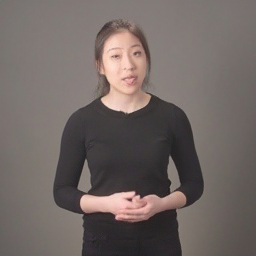}}
\subfloat{\includegraphics[width = 0.22\textwidth]{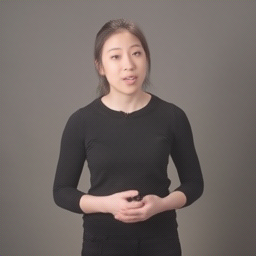}}
\subfloat{\includegraphics[width = 0.22\textwidth]{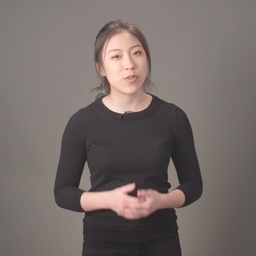}}\\[-2.5ex]
\end{center}
\caption{SeqPix2Pix synthesis results comparisons for constraint variations given the same DensePose figures in Figure~\ref{fig:VAE exmample}.}
\label{fig:GAN exmample}
\end{figure*}

To qualitatively demonstrate how each added constraint helps to converge to perceptually better local optima, in Figure~\ref{fig:GAN exmample}, we show examples comparing the SeqPix2Pix results with two baselines (i.e., Baseline 1 and Baseline 2) for different styles of DensePose frames. Compared to the baselines, the SeqPix2Pix model does a very good job at generating visually pleasing results. 

\begin{itemize}
    \item \textbf{Baseline 1} uses only a modified GAN loss \(L_{t}(G,D,P)\) by setting \((\lambda_{0},\lambda_{1},\lambda_{2})=(0.0,0.0,0.0)\). This set struggles the most with eye placement and artifact looking pixels around her mouth. 
    \item \textbf{Baseline 2} adds a structural (perceptual) constraint to \textbf{Baseline 1} by setting \((\lambda_{0},\lambda_{1},\lambda_{2})=(0.05,0.0,0.0)\). The opening and closing frames are almost entirely perfect, but the middle three frames still struggle in the eye and mouth regions. 
    \item \textbf{SeqPix2Pix} includes additional temporal consistency constraints by setting \((\lambda_{0},\lambda_{1},\lambda_{2})=(0.05,10.0,10.0)\). This set fixes the eye and mouth incompletely, but slightly blurs the instructor's teeth. Because this occurred during fast motion, the teeth blurriness is not very noticeable, especially if it was playing, rather than a still image. Overall, the perceptual improvement from \textbf{Baseline 1} set to the \textbf{SeqPix2Pix} set is significant. In particular, the instructor's eye alignment is improved greatly and mouth alignment as well.
\end{itemize}

We also quantitatively compare those three models by measuring traditional metrics MSE, PSNR, and SSIM~\cite{Wang04}. We evaluate all models on two test datasets (Test Set 1 of 4,641 figures and Test Set 2 of 3368 figures) and show results in Table~\ref{table: Qualitative results}. MSE, PSNR and SSIM rely on low-level differences between pixels and operate under the assumption of additive Gaussian noise, which may be invalid for quantitative assessments of the generated images. We therefore emphasize that the goal of these evaluations is to showcase qualitative differences between models trained with and without structural (perceptual) and/or temporal consistency constraints. For example, the SeqPix2Pix model does a very good job at generating realistic and natural faces compared to other baselines while achieving lowest SSIM scores.  

\subsubsection*{Full Video Demos, Limitations, and Future Work}
\label{sssec:full video demos}
With an instructor's two completely different audio narrations (in both lengths and contents), video lectures are produced and are available at \cite{Demo_new3:2019:Online} and \cite{Demo_new2:2019:Online} respectively. Overall, the proposed Lumi\`ereNet model produces very convincing lecture video results. The hand and body gestures are smooth. The body and hair look very realistic and natural. The hands look good, but on further examination the lines between fingers look blurry and reveal the frames as a fake. The most noticeable flaw is in the eyes. Sometimes the eyes are looking in different directions or look uneven upon close attention. While the opening and closing of lips is almost perfect sync with the narrations, finer movement details are reduced in certain time periods. We see these shortcomings come partly from the lack of those fine details in the DensePose estimator. Combining with explicit modeling of them (e.g., face keypoints from OpenPose) might enable better synthesis of such details. Moreover, to have more diverse gesture results, we think designing more informative latent codes spaces (e.g., \cite{Zhao19}) would be beneficial. 

\section{Conclusion}
\label{sec:conclusion}
In this paper, we have proposed a simple, modular, and fully neural network-based Lumi\`ereNet which produces an instructor's full pose lecture video given the audio narration input, which has not been addressed before from deep learning perspective as far as we know. Our new framework is capable of creating convincing full-pose video from arbitrary length of audio effectively. Encouraged by this result, many future directions are feasible to explore. One potential direction is to look into a latent embedding space of many instructors' video footage. Given a personalized compact latent code and a few videos of a new instructor, the system would start producing new videos after quick training. We hope that our results will catalyze new developments of deep learning technologies for commercial video content production.  

\section{Acknowledgements}
During the course of this work, the discussions and actual video shootings held with Juno Lee, Justin Lai and Larry Madrigal proved stimulating and helpful. The author also would like to thank Heidi Lim for considerable encouragements and several valuable comments on the synthesized samples. The suggestions of Stephen Welch to improve the the manuscripts also gratefully acknowledged. 

{\small
\bibliographystyle{ieee}
\bibliography{iccvbib}
}

\end{document}